\documentclass[letterpaper]{article} 
\usepackage{aaai25}  

\usepackage{times}  
\usepackage{helvet}  
\usepackage{courier}  
\usepackage[hyphens]{url}  
\usepackage{graphicx} 
\usepackage{natbib}  
\usepackage{caption} 
\usepackage{algorithm}
\usepackage{algorithmic}

\usepackage{amsmath,amsfonts}

\usepackage{amssymb}

\usepackage{textcomp}

\def\BibTeX{{\rm B\kern-.05em{\sc i\kern-.025em b}\kern-.08em
		T\kern-.1667em\lower.7ex\hbox{E}\kern-.125emX}}

\usepackage{diagbox}

\usepackage{booktabs}
\usepackage{tabularx}
\usepackage{soul}
\usepackage{array}

\usepackage{algorithmic}
\usepackage{algorithm}
\usepackage{amsmath}

\newcommand\scalemath[2]{\scalebox{#1}{\mbox{\ensuremath{\displaystyle #2}}}}

\usepackage{graphicx}
\usepackage{subfig}
\usepackage{blindtext}
\usepackage{adjustbox}
\usepackage{multirow}
\usepackage{color}
\usepackage{booktabs}
\usepackage{tabularx}
\usepackage{colortbl}

\usepackage{bbding}

\usepackage{tikz}
\usepackage{listings}
\usepackage{etoolbox}
\usepackage{subfig}
\usepackage{url}
\usepackage{tabularx}
\usepackage{array}

\definecolor{Gray}{gray}{0.9}

\newcommand{\circled}[2][]{\tikz[baseline=(char.base)]
	{\node[shape = circle, draw, inner sep = 1pt]
		(char) {\phantom{\ifblank{#1}{#2}{#1}}};%
		\node at (char.center) {\makebox[0pt][c]{#2}};}}
\robustify{\circled}

\newcommand{\SystemName}{\textsc{FLAYER}\xspace}

\newcommand\cparagraph[1]{\vspace{1.5mm} \noindent \textbf{#1}\xspace}

\usepackage{newfloat}
\usepackage{listings}
\DeclareCaptionStyle{ruled}{labelfont=normalfont,labelsep=colon,strut=off} 
\floatstyle{ruled}
\newfloat{listing}{tb}{lst}{}
\floatname{listing}{Listing}

\title{Optimizing Personalized Federated Learning through Adaptive Layer-Wise Learning}
\author {
    Weihang Chen \textsuperscript{\rm 1},
    Cheng Yang \textsuperscript{\rm 1},
    Jie Ren\textsuperscript{\rm 1},
    Zhiqiang Li\textsuperscript{\rm 1},
    Zheng Wang\textsuperscript{\rm 2}
}
\affiliations {
    \textsuperscript{\rm 1}Shaanxi Normal University, Xian, China\\
        \textsuperscript{\rm 2}University of Leeds, Leeds, United Kingdom\\
    renjie@snnu.edu.cn
}

\usepackage{bibentry}

\begin{document}

\maketitle

\begin{abstract}
Real-life deployment of federated Learning (FL) often faces non-IID data, which leads to poor accuracy and slow convergence. Personalized FL (pFL) tackles these issues by tailoring local models to individual data sources and using weighted aggregation methods for client-specific learning. However, existing pFL methods often fail to provide each local model with global knowledge on demand while maintaining low computational overhead.  Additionally, local models tend to over-personalize their data during the training process, potentially dropping previously acquired global information.
We propose \SystemName, a novel layer-wise learning method for pFL that optimizes local model personalization performance. \SystemName considers the different roles and learning abilities of neural network layers of individual local models. It incorporates global information for each local model as needed to initialize the local model cost-effectively. It then dynamically adjusts learning rates for each layer during local training, optimizing the personalized learning process for each local model while preserving global knowledge. Additionally, to enhance global representation in pFL, \SystemName selectively uploads parameters for global aggregation in a layer-wise manner. We evaluate \SystemName on four representative datasets in computer vision and natural language processing domains. Compared to six state-of-the-art pFL methods, \SystemName improves the inference accuracy, on average, by 5.4\% (up to 14.29\%).

\end{abstract}

\vspace{-6mm}
\section{Introduction}
\vspace{-1mm}

Federated Learning (FL) enables collaborative model training across diverse, decentralized data sources while preserving the confidentiality and integrity of each dataset.  It is widely used in mobile applications like private face recognition~\cite{niu2022federated}, predictive text, speech recognition, and image annotation~\cite{song2022flair}. However, data from mobile devices frequently exhibits non-IID (non-independent and identically distributed) characteristics due to variations in user behavior, device types, or regional differences~\cite{zhu2021federated}. This data heterogeneity poses significant challenges for typical FL algorithms, as the trained global model may struggle to adapt to the specific needs of individual clients, resulting in poor inference performance and slow convergence.

    Personalized Federated Learning (pFL) addresses the non-IID data challenge through client-specific learning~\cite{tan2022towards}, typically using weighted aggregation methods to customize model updates for individual clients. Customization can be achieved through various strategies, including model-wise~\cite{luo2022adapt}, layer-wise~\cite{ma2022layerwised}, or element-wise~\cite{zhang2023fedala} approaches.

Model-wise aggregation methods like APPLE~\cite{luo2022adapt}, FedAMP~\cite{huang2021personalized}, and Ditto~\cite{li2021ditto} aggregate model parameters across multiple clients, where a client downloads models from others and aggregates them locally using learned weights at the model level. This process captures comprehensive global knowledge by integrating diverse information from all participating clients. However, it can not reflect finer-grained differences in individual clients' data, potentially affecting the effectiveness of model personalization.
Additionally, it incurs significant computational costs to calculate which client model can benefit local performance, potentially limiting scalability.
Layer-wise methods perform local aggregation in layer units. Here, a ``layer unit” can be a single neural network layer or a block comprising multiple layers. Examples of layer-wise methods include FedPer~\cite{arivazhagan2019federated}, FedRep~\cite{collins2021exploiting} and pFedLA~\cite{ma2022layerwised}. These methods allow for more targeted adaptation of different parts of the network. For example, FedRep uses the global model to construct the lower layers (i.e., layers toward the input layer, also termed as base layers) of each local model and the higher layers (i.e., layers toward the model’s output layer, also termed as head layers) are built solely on local data for personalization. This approach ensures that the local model learns global representations from all clients while using the head layers to learn local representations, balancing global and local learning. Element-wise aggregation, such as FedALA~\cite{zhang2023fedala}, leverages the global model to construct the base layers, while learning an aggregation weight for each parameter in the head from both local and global models. However, these weights are set before the training stage and remain largely unchanged, potentially limiting the model's adaptability to evolving data patterns over time. Moreover, despite offering finer control, element-wise aggregation can considerably increase computational costs due to the need to compute and maintain individual weights for each parameter. Table~\ref{tab:moti1} compares the three aforementioned pFL methods with a standard FL, FedAvg~\cite{mcmahan2017communication}. In this experiment, we consider 20 clients to collaboratively learn their personalized models on the CIFAR-100 dataset~\cite{krizhevsky2009learning} using ResNet-18~\cite{he2016deep}. We simulate the heterogeneous settings by using the Dirichlet distribution $Dir(0.1)$~\cite{lin2020ensemble}. All the experiments are conducted on a single NVIDIA RTX A5000 GPU. Among the methods compared, APPLE shows a moderate improvement in accuracy over FedAvg but incurs the highest training cost among the compared methods. Conversely, FedPer introduces personalization layers (head layers) into the FL, using local data to train these layers without uploading them to the server for aggregation, thereby speeding up convergence and significantly improving inference accuracy compared to FedAvg. Building on FedPer, FedALA incorporates local and global data within each element in head layers, delivering the highest inference accuracy. However, FedALA requires learning aggregation weights on an element-wise basis, which can lead to increased computational costs. Moreover, all three pFL methods use a constant learning rate across all clients during training, without considering the differing learning needs of various layers for each client. This may lead to over-personalization for each client, resulting in the loss of previously aggregated global information.

\begin{table}[t]      
	\scriptsize
	\begin{tabular}{@{}lrrrr@{}}	

        \toprule
        Method & $\#$Iter.  & Total time (s) & Time (s) /iter.  & Acc. ($\%$)   \\
         \midrule
           FedAvg  &  181   &5430    &  30   & 37.08 \\
            APPLE (model-wise)  &  25    &6000    &  240  & 57.29 \\
           FedPer (layer-wise)  &  184    &5704   &  31   & 54.26 \\
           FedALA (element-wise) &  76    & 2660   &  35   & 58.65 \\
        \bottomrule

    \end{tabular}
\vspace{-3mm}
         	\caption{The computation cost (includes \# training iterations until convergence, total training time, and training time in each iteration) and inference accuracy (\%) on CIFAR-100 using ResNet-18.}
       
             \label{tab:moti1}
                      \vspace{-7mm}
\end{table}

\begin{figure*}[t!]
	\begin{center}
		\includegraphics[width=0.9\textwidth]{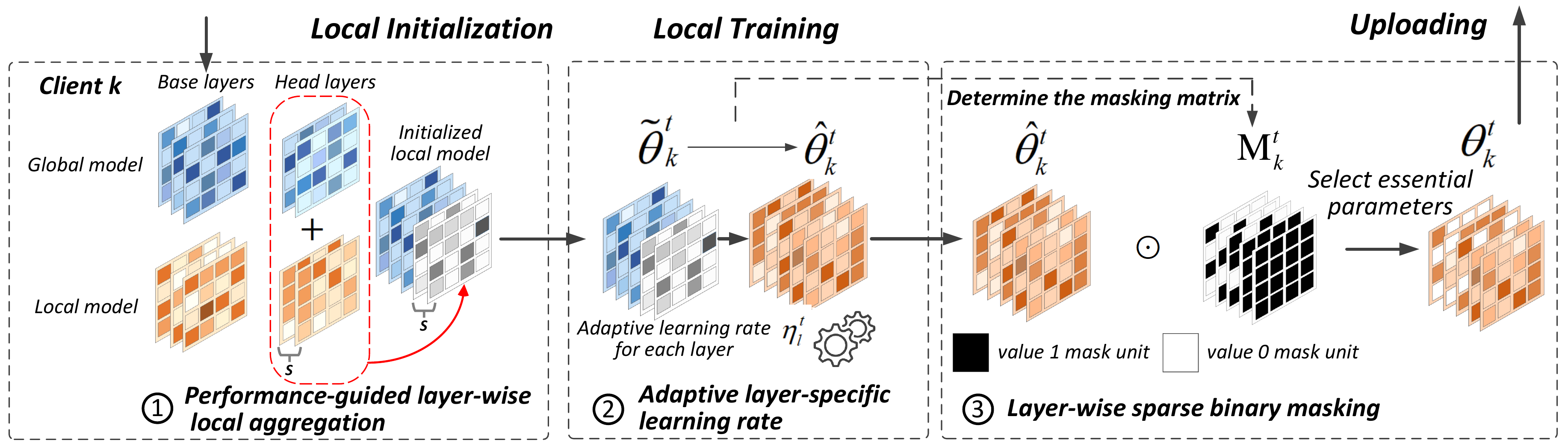}
	\end{center}
 	\vspace{-4mm}
	\caption{The local learning process of \SystemName on $k$-th client during the $t$-th iteration. In the local initialization stage, \SystemName aggregates both local and global head layers based on the local model's performance from the previous iteration. The initialized local model is then trained on the local data, using an adaptive learning rate for each layer. Based on the parameter changes before and after local training, \SystemName constructs a masking matrix to identify and select essential parameters, with different proportions per layer, for updating the global model.}
	 \vspace{-5mm}
	\label{fig:overview}
\end{figure*}

To effectively incorporate local and global information across all network layers on each client, we introduce \SystemName, a new layer-wise optimization for pFL. \SystemName operates throughout the local learning process: local model initialization, training, and parameter uploading. Specifically, during the local model initialization stage, we aggregate local and global information in the head layers on a layer-wise basis, with aggregation weights guided by the local model's inference accuracy, which can be easily accessed on each client with negligible cost. This allows for dynamic adjustment of the learning contributions from local and global models based on local performance.  In the local model training stage, \SystemName applies a layer-wise adaptive learning rate scheme based on each layer's position and gradient. This enables each layer to effectively learn from the local dataset after initialization and helps mitigate the issue of vanishing gradient during training. For parameter uploading, \SystemName implements a layer-wise masking strategy to select essential parameters from each client for global aggregation. This ensures that the global averaging process retains crucial base features, enhancing the overall effectiveness of global learning.

We evaluate \SystemName by applying it to both image~\cite{krizhevsky2009learning,chrabaszcz2017downsampled} and text~\cite{zhang2015character} classification tasks using four widely adopted benchmarks. The results show that \SystemName outperforms six other pFL methods ~\cite{luo2022adapt,li2021ditto,huang2021personalized, arivazhagan2019federated,collins2021exploiting, zhang2023fedala} in inference accuracy and computational cost. This paper makes the following contributions:

\begin{itemize}

\item A new performance-guided layer-wise aggregation method allows clients to dynamically incorporate both local and global information in a cost-effective manner;
\item A new layer-specific adaptive learning rate scheme for pFL to steer the personalization and speed up convergence;
\item A new layer-wise masking technique for selectively uploading essential parameters to the central server to improve global representation.

\end{itemize}


\vspace{-4mm}
\section{Background and Overview}
\vspace{-1mm}

\subsection{Problem Definition}
\vspace{-1mm}
pFL learns personalized models cooperatively among clients. Consider a scenario where we have $n$ clients, and each client processes their distinct private training data denoted as $D_1, D_2, \ldots, D_N$, which has different data classes and sizes. These datasets exhibit heterogeneity, characterized by non-IID (non-independently and identically distributed)~\cite{zhao2018federated}.
The goal of pFL can be defined as: 
\vspace{-3mm}
\begin{equation}
	\{\theta_1, \theta_2, ..., \theta_n\} = \arg\min_{\theta} \sum_{k=1}^n \frac{m_k}{M}L_k(\theta_k)
\end{equation}
\vspace{-3mm}

where $\theta_k$ is the model parameters of client $k$, $m_k$ is the data size of $D_k$, $M$ is the whole data size of all clients, $L_k(\theta_k)$ is the loss function of client $k$.

\vspace{-2mm}
\subsection{Overview of \SystemName\label{sec:selector}}

Figure~\ref{fig:overview} depicts the workflow of \SystemName during the training of the $k$-th client in the $t$-th iteration. Initially, the global model is downloaded to each client for local aggregation, \SystemName implements an adaptive aggregation strategy, dynamically adjusting the aggregation weights between the global and local models based on the inference accuracy of the local model. 
This allows models, whether local or global, with higher performance to have a greater influence on the initialization of the local head layer.
During the local training phase, \SystemName leverages a layer-wise adaptive learning rate scheme. This strategy dynamically adjusts the learning rates according to each layer's position within the network and the gradient of that layer. By optimizing the learning step for each layer, \SystemName enhances local model personalization performance and accelerates convergence. In the final phase of updating the model to the central server, \SystemName employs a layer-wise masking strategy. This selective approach uploads only the essential parameters from each local model, preventing the global averaging process from diluting the crucial information captured by the local models, thus enhancing the global representation. Algorithm~\ref{algo:overall} presents the overall FL process in \SystemName.

\vspace{-1mm}
\section{Methodology}
\vspace{-2mm}
\subsection{Performance-guided Layer-wise Local Aggregation\label{sec:ala}}
\vspace{-2mm}

Building on previous findings~\cite{yosinski2014transferable,zhu2021data} that base layers of DNN capture generalized features while head layers encode task-specific features, we introduce a differentiated update strategy for these layers. In our approach, during the local model initialization stage, the base layers of each client's model are directly updated using parameters from the global model. This ensures the consistent refinement of generalized features across all clients. For the head layers, we use a dynamic aggregation method to integrate both global and local parameters. This integration is tailored based on the performance of the $k$-th client's model on its local dataset $D_k$ from the previous iteration. The process is defined as follows:
\begin{gather}
\scalemath{0.8}{\tilde\theta_k^t := [\theta_g^{(1:L-s,t-1)}, A_{k,l}^{t-1}\odot \theta_k^{(L-s+1:L,t-1)} + A_{k,g}^{t-1}\odot \theta_g^{(L-s+1:L,t-1)}} ]\nonumber\\
\scalemath{0.8}{\text{s.t.}\quad  A_{k,l} +  A_{k,g} = 1\label{equ:la}} 
\end{gather}
\vspace{-5mm}

Here, $\theta_k$ is the local model parameter matrix of the $k$-th client,
 $\tilde\theta_k$ denotes the local model parameter matrix of the $k$-th client after initialization, $L$ is the total number of layers, $s$ is the number of layers in the head for personalization, and $\theta_g^{(1:L-s,t-1)} $ represents the lower $L-s$ layers in the base part of the global model at iteration $t-1$, which are used to update the base layers in the local model, and all the clients share the same base layers. $\theta_k^{(L-s+1:L,t-1)}$ denotes the head layers of the $k$-th local model. We aggregate the head layers from the local and global models to initialize the head layers for the local model. The aggregation weights
$A_{k,g}$ and $A_{k,l}$  control the influence of global and local parameters, respectively. At iteration ${t-1}$, the local model inference accuracy on dataset $D_k$ sets the local weight $A_{k,l}^{t-1}$, with $(1- A_{k,l})$ adjusting for global influence. A lower local accuracy increases reliance on the global model through $A_{k,g}$,  providing stability in early training phases. As the accuracy of the local model increases, the initialization of the head layer becomes more dependent on the local model, thereby incorporating more localized and personalized information.

\vspace{-2mm}
\subsection{Adaptive Layer-specific Learning Rate\label{sec:alr}}
After the model initialization, the local model trains on its local dataset $D_k$:
\begin{gather}
\hat\theta_k^t := \tilde\theta_k^t - \eta \nabla_{\tilde\theta_k} \mathcal{L}_k(\tilde\theta_k^t, D_k)
\end{gather}

$\eta$ represents the learning rate, $\nabla_{\tilde\theta_k} \mathcal{L}_k(\tilde\theta_k^t,D_k)$ is the gradient of the loss function $\mathcal{L}_k$ with respect to the parameters $\tilde\theta_k$ evaluated using local dataset $D_k$.

The existing pFL methods~\cite{luo2022adapt,li2021ditto,huang2021personalized, arivazhagan2019federated,collins2021exploiting, zhang2023fedala} typically employ a fixed learning rate. However, in the context of FL with non-IID data, we observe that the learning rate is a critical hyperparameter that significantly impacts both the performance and convergence speed of local models (see Section \textit{Ablation}). Previous work~\cite{singh2015layer} pioneered layer-wise learning rate adjustments, primarily to mitigate the vanishing gradient issue in the lower layers of DNNs within a non-distributed training context. However, this approach is not well-suited for the pFL context. Inspired by~\cite{luo2021no} and our observation (see Section \textit{Layer Similarity}), we note that the first layer among all local models, showing the highest similarity, captures universal features and thus requires a smaller learning rate with more gradual adjustments. In contrast, deeper layers, exhibiting greater divergence, deal with more complex, client-specific features and benefit from larger learning steps, which aids local model personalization. 
 Building on these insights, we implement an adaptive learning rate scheme for pFL that integrates layer positional information with the corresponding gradient:
\begin{equation}
\eta^{(i,t)} = \eta \left(1 + \log\left(1 + \frac{1}{\|g^{(i,t)}\|_2}\right)\times\frac{i}{L}\right)
\end{equation}
where:
\begin{itemize}
    \item $\eta$ is the base learning rate.
    \item $g^{(i,t)}$ represents the gradient vector of the $i$-th layer at iteration $t$.
    \item $\|g^{(i,t)}\|_2$ is the L2 norm (Euclidean norm) of the gradient.
    \item $L$ is the total number of layers.
\end{itemize}

\vspace{-3mm}
\subsection{Layer-wise Sparse Binary Masking\label{sec:lbm}}
In non-IID settings, averaging updated client parameters on the server side can dilute important information during aggregation. To address this, we propose a layer-wise binary masking scheme for server aggregation, aimed at preserving critical information and ensuring a high-quality global representation. The core idea is to selectively upload parameters from each layer based on their significance and layer position, distinguishing between general features in early layers and more complex, client-specific features in deeper layers~\cite{luo2021no}. In detail, our strategy prioritizes uploading parameters with high significance, typically those with greater changes in early layers, as these are likely to capture essential patterns and features common to all client datasets, thereby enhancing the model's generalization ability. For deeper layers, we employ a more inclusive approach, uploading a greater proportion of parameters to capture a wide range of client-specific details and complex features essential for the model's performance on localized tasks. The proportion of parameters uploaded from each layer, denoted as \(UP^i\) for layer \(i\), is calculated based on the layer's position within the network architecture using the following formula:
\vspace{-2mm}
 \begin{equation}
UP^i := \min(\max(\frac{i}{L}, 0.1), 1)
 \end{equation}
\vspace{-3mm}

where $UP^i$ is constrained to be at least 0.1 to ensure that every layer contributes to the global model aggregation, but not more than 1, reflecting a full update contribution.

To identify and select significant weights for sharing, we calculate the absolute weight fluctuation value of the local model within each layer after local training:

\vspace{-2mm}
 \begin{equation}
 	\Delta{\theta}_k^{(i,t)}  := |\hat{\theta}_k^{(i,t)} - \tilde\theta_k^{(i,t)}|
\end{equation}

Then, to focus on parameters that have undergone notable changes, we identify the top \(UP^i\) parameters from each layer
 \(i\) in $k$-th client model based on their fluctuation values after the \(t\)-th training round:

\vspace{-3mm}
\begin{equation}
    S_k^{(i,t)} := \textit{top\_percent}\left(\Delta{\theta}_k^{(i,t)}, UP^i\right)
\end{equation}

To manage which parameters are uploaded from each layer \(i\) of a local model on client \(k\), we use a binary mask matrix, \(M_k^t\),  which has the same dimensions as the parameter matrix \(\hat{\theta}_k^{t}\). Initially, all elements of this matrix are initialized to one. Each item \(m_{j,k}^{(i,t)}\) in \(M_k^{(i,t)}\) is determined by whether the corresponding parameter \(\theta_{j,k}^{(i,t)}\) in \(\Delta{\theta}_k^{(i,t)}\) belongs to the subset of parameters with the highest weight changes, \(S_k^{(i,t)}\). This setup uses the following rule:

\vspace{-2mm}
\begin{equation}
    m_{j,k}^{(i,t)} =
    \begin{cases}
    1, & \text{if } \theta_{j,k}^{(i,t)} \in S_k^{(i,t)} \\
    0, & \text{otherwise}
    \end{cases}
\end{equation}
\vspace{-2mm}

Finally, we obtain the essential parameter $\theta_k^t$ required for uploading by multiplying the $\hat\theta_k^t$ with the binary mask $M_k^t$.

\begin{equation}
	\theta_k^t := \hat\theta_k^t \odot M_k^t
\end{equation}

\captionsetup[algorithm]{name=Algorithm}
\begin{algorithm}[tb]
\caption{\SystemName}
\label{algo:overall}
\textbf{Input}: $N$ clients, $\rho$: client joining ratio, $L$: loss function,
		$\Theta_g^0$: initial global model, $\eta$: base local learning rate, 
		$s$: the hyperparameter of \SystemName.\\
\textbf{Output}: Well-performing local models $\tilde\Theta_1, \ldots, \tilde\Theta_N$

\begin{algorithmic}[1]
		\STATE Server sends $\Theta_g^0$ to all clients to initialize local models.
		\FOR{iteration $t=1, \ldots, T$ }
		\STATE Server samples a subset $C^t$ of clients according to $\rho$.
		\STATE Server sends $\Theta_g^{t-1}$ to $\left|C^t\right|$ clients.
		\FOR{Client $k \in C^t$ in parallel}
		\STATE Client $k$ initializes local model $\tilde{\Theta}_k^t$ by Equation (2).
		\STATE Client $k$ obtains $\hat\Theta_k^t$ by Equation (3) - (4). 
        \STATE$\qquad\qquad\qquad\qquad\qquad\quad\triangleright$ Local model training
		\STATE Client $k$ obtains masked $\Theta_k^t$ by Equation (5) - (9).
		\STATE Client $k$ sends $\Theta_k^t$ to the server. $\qquad\triangleright$ Uploading
		\ENDFOR
		\STATE Server obtains $\Theta_g^t$ by $\Theta_g^t \leftarrow \sum_{k \in \mathrm{C}^t} \frac{n_k}{\sum_{j \in \mathrm{C}^t} n_j} \Theta_k^t$.
		\ENDFOR\\
		\STATE \textbf{return} $\tilde\Theta_1, \ldots, \tilde\Theta_N$

\end{algorithmic}
\end{algorithm}


Adopting selective weight sharing, \SystemName enhances the global model representation. Our approach differs from FedMask~\cite{li2021fedmask}, which also achieves personalization using a heterogeneous binary mask with a small overhead. However, FedMask does not consider the unique characteristics of different layers, failing to capture layer-specific information. Moreover, FedMask's binary parameter aggregation is insufficient for complex tasks, such as CIFAR-100. In our approach, early layers, which capture universal features, are updated only with the most critical changes during server aggregation, preserving a robust foundation for all clients and preventing the dilution of essential base features. Conversely, deeper layers, which capture complex and client-specific features, receive updates from a larger proportion of parameters. This ensures the global model incorporates a diverse set of features, enhancing its generalization ability.

\vspace{-3mm}
\section{Evaluation Setup\label{sec:setup}}

\begin{table*}[!t]
\centering

  \footnotesize
    \begin{tabular}{@{}l|rrr|rrr|c@{}}
	  \toprule
		&\multicolumn{3}{c|}{CNN} &\multicolumn{3}{c|}{ResNet-18}   &\multicolumn{1}{c}{fastText}\\
		\midrule
		Method &CIFAR-10 &CIFAR-100 &Tiny-ImageNet&CIFAR-10&CIFAR-100 &Tiny-ImageNet & AG News\\ 
		\midrule			
		FedAvg & 59.16$\pm$0.56 &33.08$\pm$0.61 &18.86$\pm$0.29 &86.95$\pm$0.39 &37.08$\pm$0.43 &20.32$\pm$0.20 &80.12$\pm$0.31\\
  		APPLE (model) & 89.60$\pm$0.16 &54.45$\pm$0.24 &39.42$\pm$0.49 & 89.78$\pm$0.19 &57.29$\pm$0.30 &43.26$\pm$0.55 &95.37$\pm$0.23\\
      	Ditto (model)&89.48$\pm$0.04 &47.68$\pm$0.59 &33.89$\pm$0.08 & 88.70$\pm$0.18 &48.46$\pm$0.89 &36.37$\pm$0.52 &94.66$\pm$0.18\\
    	FedAMP (model)&89.31$\pm$0.17 &47.77$\pm$0.46 &33.82$\pm$0.33 & 88.52$\pm$0.22 &48.75$\pm$0.49 &35.83$\pm$0.25 &94.02$\pm$0.11\\
		FedPer (layer)& 89.55$\pm$0.28 &49.15$\pm$0.57 &39.61$\pm$0.24 &89.20$\pm$0.21 &54.26$\pm$0.43 &42.38$\pm$0.55 &95.07$\pm$0.16\\
		FedRep (layer)&90.62$\pm$0.18 &51.45$\pm$0.31 &41.79$\pm$0.52 & 90.29$\pm$0.29 &53.94$\pm$0.40 &45.98$\pm$0.72 &96.47$\pm$0.15\\
		FedALA (element)&90.84$\pm$0.09 &56.98$\pm$0.18 &45.10$\pm$0.25 & 91.30$\pm$0.35 &58.65$\pm$0.26 &49.09$\pm$0.89 &96.58$\pm$0.10\\
		\SystemName &  \textbf{91.66$\pm$0.05} & \textbf{60.50$\pm$0.33} & \textbf{45.88$\pm$0.29} &  \textbf{91.68$\pm$0.21} & \textbf{60.68$\pm$0.42} & \textbf{50.12$\pm$0.36} &
        \textbf{98.27$\pm$0.22}\\
	  \bottomrule
	\end{tabular}
     \vspace{-3mm}
    	\caption{The average inference accuracy ($\%$) across all clients on CIFAR-10, CIFAR-100, Tiny-ImageNet and AG News.}	

\label{tab:overall-acc}
		 \vspace{-4mm}
\end{table*}

\subsection{Platforms and Workloads}\label{sec:paw}
\vspace{-1mm}
To evaluate the performance of \SystemName,  we use a four-layer CNN~\cite{mcmahan2017communication} and ResNet-18~\cite{he2016deep} for CV tasks, training them on three benchmark datasets: CIFAR-10, CIFAR-100~\cite{krizhevsky2009learning}, and Tiny-ImageNet~\cite{chrabaszcz2017downsampled}. For the NLP task, we train fastText~\cite{joulin2017bag} on the AG News dataset~\cite{zhang2015character}.
We use the Dirichlet distribution $Dir(\beta)$ with $\beta = 0.1$~\cite{lin2020ensemble,wang2020tackling} to model a high level of heterogeneity across client data. Following FedAvg, we use a batch size of 10 and a single epoch of local model training per iteration. We execute the training process five times for each task and calculate the geometric mean of training latency and inference accuracy until convergence. Our experiments consider 20 clients. The number of layers in the head for CNN, ResNet-18, and fastText is 1, 2 and 1, respectively. Following FedALA, we set a base learning rate of 0.1 for ResNet-18 and fastText and 0.005 for CNN during local training. All experiments were conducted on a multi-core server with a 24-core 5.7GHz Intel i9-12900K CPU and an NVIDIA RTX A5000 GPU with 24 GB of GPU memory.


\vspace{-3mm}
\subsection{Competitive Baselines}
We compare \SystemName with six other pFL methods alongside  FedAvg, including model-wise aggregation methods APPLE, Ditto and FedAMP, layer-wise aggregation methods FedPer and FedRep, and element-wise FedALA on four popular benchmark datasets in inference accuracy. In addition, we also evaluate the performance of \SystemName in terms of the computation cost, hyperparameter, layer similarity, data heterogeneity, scalability, and applicability.

\renewcommand{\arraystretch}{0.8}
\begin{table}[t!]
\centering

     \scriptsize
    \begin{tabular}{@{}l|l|>{\raggedleft\arraybackslash}p{0.5cm}r|>{\raggedleft\arraybackslash}p{0.5cm}r@{}}
		\toprule
		\multicolumn{1}{c|}{} & \multicolumn{1}{l|}{Model} & \multicolumn{2}{c|}{CNN} & \multicolumn{2}{c}{ResNet-18}\\
        \midrule
         Dataset & Method &   $\#$Iter. 
& Total time (s)   
&  $\#$Iter. 
& Total time (s)  
\\
		\midrule			
		\multirow{6}{*}{CIFAR-10} & FedAvg 	& 157& 
1256& 179& 

5191 
\\
           & APPLE	& 190& 
6650& 130& 

31070\\
             & Ditto & 51& 
1071& 
172& 
11696 
\\
               & FedAMP 	& 47& 
$^\#$\textbf{517}& 
191& 
7067 
\\
		  & FedPer & 156& 
1248& 
183& 
5307 
\\
    	  & FedRep 	& 169& 
2028& 
185& 
6845 
\\
            & FedALA& 152& 
1520& 
133& 
5187 
\\
            &\textbf{\SystemName}	& 78& 
858& 
53& 
$^\#$\textbf{2067} 
\\
		\midrule            
		\multirow{6}{*}{CIFAR-100} & FedAvg	& 180& 
1620& 181& 

5430 
\\
           & APPLE	& 195& 
6825& 25& 

6000\\
             & Ditto	& 57& 
1254& 
101& 
6868 
\\
               & FedAMP 	& 61& 
671& 
173& 
6401 
\\
		  & FedPer 	& 101& 
909& 
184& 
5704 
\\
    	  & FedRep	& 69& 
828& 
179& 
6802 
\\
            & FedALA	& 120& 
1200& 
76& 
2660 
\\
            & \textbf{\SystemName} 	& 27& 
$^\#$\textbf{324}& 58& 

$^\#$\textbf{2378} 
\\
		\midrule            
		\multirow{6}{*}{Tiny-ImageNet} & FedAvg 	& 48& 
2016& 74& 

5920 
\\
           & APPLE	& 69& 
9867& 37& 

17427\\
             & Ditto	& 35& 
3150& 
174& 
29754 
\\
               & FedAMP	& 28& 
1316& 
84& 
7392 
\\
		  & FedPer 	& 31& 
1302& 
78& 
6240 
\\
    	  & FedRep 	& 39& 
1794& 
115& 
10350 
\\
            & FedALA	& 64& 
2944& 
48& 
4368 
\\

            & \textbf{\SystemName} 	& 16& $^\#$\textbf{896}& 18& $^\#$\textbf{1782}\\
		\bottomrule
	\end{tabular}
 \vspace{-3mm}
  	\caption{The average computation cost for CV tasks.}	\label{tab:overall-cost}
   		\vspace{-5mm}
\end{table}

\vspace{-3mm}
\section{Experimental Results\label{sec:result}}

\subsection{Overall Performance}
\vspace{-2mm}

\cparagraph{Inference accuracy.}
Table~\ref{tab:overall-acc} compares the inference accuracy of \SystemName with six other SOTA pFL methods in CV and NLP domains with $Dir(0.1)$. APPLE gives the highest accuracy in the model-wise category, but with a high computation cost. FedPer uses a simple local aggregation strategy, utilizing global base layers and local head layers to initialize the local model, improving accuracy by an average of 18.1\% over FedAvg. FedRep further enhances this by separately training head and base layers, boosting accuracy by 19.7\% over FedAvg.
Building on FedPer, FedALA incorporates global information into the local head initialization, achieving a 22.7\% improvement in accuracy compared to FedAvg. Previous layer-wise pFL methods recognize the different roles of base and head layers in non-IID settings and apply different strategies for integrating global and local information to initialize the local model. However, they often overlook the roles and learning capabilities of the base and head layers during the local training stage. This oversight prevents the layers from capturing local information on demand, potentially slowing down convergence speed. \SystemName achieves the highest test accuracy among all pFL methods, with a 24.17\% improvement over FedAvg. 
This is achieved by effectively incorporating global and local information for each client in a layer-wise manner during the initialization, local training, and model updating stages.

\vspace{-1mm}
\cparagraph{Computation cost.}
Table~\ref{tab:overall-cost} compares the computation cost of our approach with six other pFL methods and FedAvg, measured by the training time required for convergence. Except for CIFAR-10 with CNN, where FedAMP delivers the lowest training cost (but with poor inference accuracy), \SystemName gives the lowest computation cost across all other tasks, reducing total training cost by an average of 58.9\% (up to 80.1\%) compared to FedAvg.
Specifically, model-wise methods like APPLE and Ditto involve complex calculations leading to high overhead. FedRep trains the base and head layers separately, which incurs significant training costs. \SystemName effectively incorporates both local and global information across all layers, resulting in fewer rounds needed for convergence compared to FedALA, with an average reduction of 52.7\% in total training time.

\begin{table}[]
\centering

\scriptsize
\begin{tabular}{@{}l|rrr|rrr@{}}

\toprule

               & \multicolumn{3}{c|}{CNN} & \multicolumn{3}{c}{ResNet-18} \\ \midrule
Hyperparameter (s)    & 3    & 2   & 1   
 & 3  & 2  & 1  \\
\midrule
Accuracy (\%)     & 53.58     & 54.42     &  $^\#$\textbf{60.50}   
  & 59.80   &  $^\#$\textbf{60.68}   & 60.16   \\ \bottomrule
\end{tabular}
\vspace{-3mm}
\caption{The inference accuracy ($\%$) of \SystemName on CIFAR-100 by using CNN and ResNet-18 with various $s$.}	

\label{tab:hypers}
\vspace{-6mm}
\end{table}

\begin{table*}[!t]
\footnotesize
\centering
\begin{tabular}{@{}l|rr|rrr|rr@{}}
\toprule
  & \multicolumn{2}{c|}{\textbf{Heterogeneity}}& \multicolumn{3}{c|}{\textbf{Scalability}} & \multicolumn{2}{c}{\textbf{Applicability}} \\ \midrule
  \textbf{Methods} & \textbf{Dir(0.1)} & \textbf{Dir(0.01)} & \textbf{20 clients} & \textbf{50 clients} & \textbf{100 clients} &\textbf{Acc.} &\textbf{Imps.}     \\  \midrule
 FedAvg & 37.08$\pm$0.43 & 43.74$\pm$0.38& 37.08$\pm$0.43 & 34.56$\pm$0.25 & 33.08$\pm$0.41 & 60.68$\pm$0.42 & 24.77\\
 APPLE & 57.29$\pm$0.30 & 74.52$\pm$0.19 & 57.29$\pm$0.30 & 58.09$\pm$0.24 & 48.46$\pm$0.32 & - & - \\
 Ditto & 48.46$\pm$0.89 & 72.94$\pm$0.22 & 48.46$\pm$0.89 & 46.08$\pm$0.19 & 43.42$\pm$0.37 & 58.49$\pm$0.21 & 10.03 \\
 FedAMP & 48.75$\pm$0.49 & 73.12$\pm$0.17 & 48.75$\pm$0.49 & 46.49$\pm$0.44 & 43.74$\pm$0.20 & 60.72$\pm$0.27 & 11.97 \\
  GPFL & 51.06$\pm$0.42 & 74.59$\pm$0.21 & 51.06$\pm$0.42 & 48.30$\pm$0.29 & 44.61$\pm$0.32 & -&- \\
 FedPer & 54.26$\pm$0.43 & 73.52$\pm$0.15 & 54.26$\pm$0.43 & 51.24$\pm$0.39 & 47.67$\pm$0.36 & 63.13$\pm$0.23 & 8.87 \\
 FedRep & 53.94$\pm$0.40 & 75.08$\pm$0.18 & 53.94$\pm$0.40 & 50.10$\pm$0.30 & 45.80$\pm$0.27 & 61.33$\pm$0.17 & 7.39 \\
  FedCP & 46.72$\pm$0.38 & 69.42$\pm$0.32 & 46.72$\pm$0.38 & 42.86$\pm$0.33 & 40.19$\pm$0.24 & - & -\\
 FedALA & 58.65$\pm$0.26 & 75.24$\pm$0.11 & 58.65$\pm$0.26 & 59.46$\pm$0.23 & 58.80$\pm$0.41 & 63.55$\pm$0.58 & 4.90 \\
 \textbf{\SystemName} & $^\#$\textbf{60.68$\pm$0.42} &$^\#$\textbf{77.39$\pm$0.24} & $^\#$\textbf{60.68$\pm$0.42} & $^\#$\textbf{61.70$\pm$0.30} & $^\#$\textbf{59.96$\pm$0.39} & - & - \\
\bottomrule
\end{tabular}
\vspace{-3mm}
\caption{The inference accuracy (\%) of eight FL methods across varying levels of statistical heterogeneity and scalability, and the performance improvement (\%) when applying our approach to them using ResNet-18 on CIFAR-100.} \label{tab:comb}

\end{table*}

\vspace{-2mm}
\subsection{Evaluation on Personalization Layers\label{sec:pl}}
\vspace{-1mm}
Table~\ref{tab:hypers} shows inference accuracy for a 4-layer CNN and ResNet-18 with varying sizes (termed as \textbf{\textit{s}}) of the head layers. For ResNet-18, the highest inference accuracy is achieved with \textbf{\textit{s}} set to 2, focusing personalization on the final two layers. For the 4-layer CNN, \textbf{\textit{s}} is set to 1, with the remaining layers updated using the global model.

\vspace{-2mm}
\subsection{Layer Similarity\label{sec:sim}}
\begin{figure}
	\centering
	\subfloat[][CNN at round 1]{\includegraphics[width=0.18\textwidth]{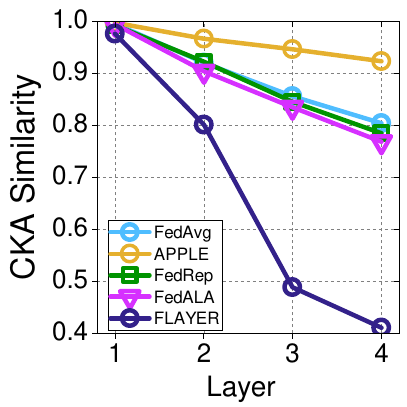}}
 \hspace{6mm}
	\subfloat[][CNN until convergence]{\includegraphics[width=0.18\textwidth]{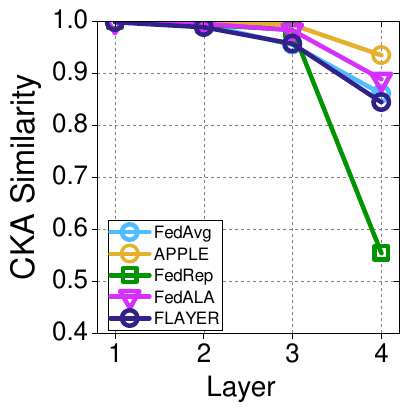}}\\
 \vspace{-4mm}
	\subfloat[ResNet-18 at round 1]{\includegraphics[width=0.18\textwidth]{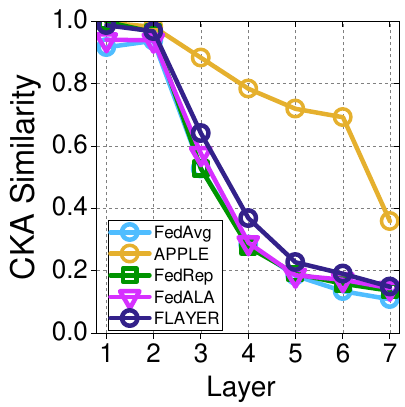}}
  \hspace{6mm}
	\subfloat[][ResNet-18 until convergence]{\includegraphics[width=0.18\textwidth]{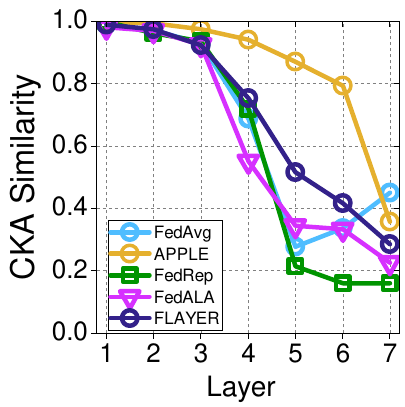}}
 \vspace{-4mm}
	\caption{The average CKA similarities of the same layers in different local models with CIFAR-10 under $Dir(0.1)$.}
	\label{fig:cka}
 \vspace{-6mm}
\end{figure}

To analyze how pFL methods perform across layers on non-IID datasets, we measure the Centered Kernel Alignment (CKA)~\cite{kornblith2019similarity} similarity of features from the same layer of different clients' models using identical test samples. This analysis helps to evaluate the balance between personalization and generalization of different pFL methods. Figure~\ref{fig:cka} presents the CKA similarity across 20 clients for methods like FedAvg, APPLE, FedRep, FedALA, and \SystemName on CIFAR-10, highlighting changes from the initial round to the training convergence. We observe that after training, the similarity of the base layers in both the CNN and ResNet-18 improves across all FL methods, indicating that the global model effectively captures common features shared by different clients. The deeper layers show lower similarity, with the head layers exhibiting the least, reflecting their focus on localized, client-specific data. Additionally, the simpler structure of the 4-layer CNN results in higher similarity across all layers compared to ResNet-18, suggesting it is less capable of capturing specialized features. \SystemName achieves moderate similarity levels in the base and head layers, suggesting that it balances well between integrating global patterns and adapting to local specifics, thereby enhancing overall model performance.

\vspace{-2mm}

\subsection{Evaluation on Data Heterogeneity}
We also evaluate the impact of statistical heterogeneity on \SystemName and other SOTA pFL methods using 20 clients. Specifically, we set two degrees of heterogeneity on CIFAR-100. The first scenario is $\beta=0.01$, where the smaller the value of $\beta$, the greater the heterogeneity of the setting. We use $\beta=0.1$ as the baseline performance. Table \ref{tab:comb} reports the performance impact under these varying degrees of heterogeneity. \SystemName consistently outperforms all other SOTA pFL methods across all heterogeneous settings, delivering the highest accuracy.

\vspace{-2mm}
\subsection{Scalability}
To evaluate the scalability of our approach, we vary the number of clients from 20 to 100 when applying ResNet-18 to CIFAR-100, setting the heterogeneity parameter $Dir(0.1)$. Table \ref{tab:comb} compares the average inference accuracy between \SystemName and other pFL methods. We can see that \SystemName consistently outperforms others across various scales of client quantity. While a decrease in accuracy across all methods is observed as the client count increases from 50 to 100, \SystemName experiences a decline of less than 2\%. In contrast, the APPLE method shows a significant drop in performance, with a 9.6\% decrease in inference accuracy in the same scenario. This underlines the efficiency of \SystemName in managing larger numbers of clients, particularly in scenarios characterized by increased scalability demands.

\vspace{-2mm}
\subsection{Applicability}
   \vspace{-1mm}
   
Our evaluation so far applied \SystemName to FedAvg. We now apply \SystemName to other FL methods to evaluate the generalization ability of our approach. Note that \SystemName does not replace the foundational architectures of an FL method. Table~\ref{tab:comb} reports the inference accuracy and improvements achieved after applying our approach to an underlying FL method. \SystemName improves the accuracy of all pFL methods, boosting the accuracy by 4.90\% to 11.97\%.

\begin{figure}[t]
	\centering

 	\subfloat[][CNN]{\includegraphics[width=0.22\textwidth]{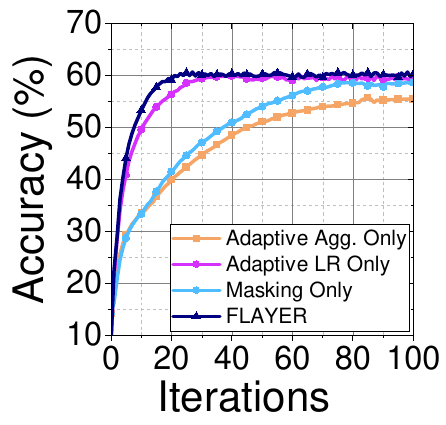}}
	\subfloat[][ResNet-18]{\includegraphics[width=0.216\textwidth]{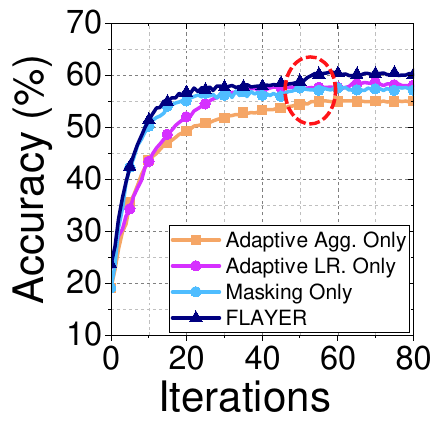}}
 \vspace{-3mm}
	\caption{The ablation study with CIFAR-100, conducted under a $Dir(0.1)$ distribution.}
	\vspace{-6mm}
	\label{fig:ablation}
\end{figure}

\vspace{-3mm}
\subsection{Ablation Study\label{sec:abl}}
   \vspace{-1mm}
Figure~\ref{fig:ablation}  presents the accuracy of three strategies in \SystemName when used alone: \textit{Adaptive Aggregation (Agg.) Only}, \textit{Adaptive Learning Rate (LR) Only}, and \textit{Masking Only.} The results show that \textit{Adaptive LR Only} achieves the highest accuracy on both ResNet-18 and CNN when trained on the CIFAR-100. For CNN, the full \SystemName exhibits a convergence speed similar to \textit{Adaptive LR Only}, suggesting that the learning rate is the most influential factor for CNN performance. While \textit{Masking Only} shows a convergence trend comparable to \SystemName on ResNet-18, it converges more slowly on CNN compared to the other two strategies. The \textit{Masking Only} benefits deeper network structures by prioritizing critical parameters involved in residual connections, thereby preserving the integrity of these connections and enhancing performance and convergence in deeper networks like ResNet-18. \textit{Adaptive Agg. Only} brings the least benefit and has the slowest convergence speed when used alone. However, its role is essential in incorporating local and global information in the head layers before training, which lays a solid foundation for the effectiveness of adaptive learning rate and masking strategies. During ResNet-18 training, accuracy improves significantly around the 50th iteration, aligning with the trend observed in \textit{Adaptive Agg. Only}. While \textit{Adaptive LR Only} is crucial for performance enhancements, particularly in CNNs, the combined approach of \textit{Adaptive Agg.}, \textit{Adaptive LR}, and \textit{Masking} within \SystemName offers a balanced and synergistic strategy that leverages the strengths of each scheme.

\vspace{-2mm}
\subsection{Discussion} 
\vspace{-2mm}

\cparagraph{Computation cost.} 
\SystemName introduces additional computational tasks for FL clients, such as calculating the L2 norm of the gradient per layer for adaptive learning rates and creating a masking matrix for critical parameters. Although these tasks incur per-iteration costs, they are offset by a reduction in overall training time. We further reduce computation costs using parallel processing and PyTorch’s optimized operations. Future plans include deploying \SystemName on real-world FL testbeds and enhancing efficiency for resource-limited devices through advanced caching and hierarchical FL strategies.

\cparagraph{Application scenarios.} 
\SystemName supports mobile applications such as predictive text and image annotation by training personalized models directly on devices, ensuring privacy and relevance to user preferences. The system optimizes model performance through adaptive learning and reduces battery impact by conducting training during charging periods. Future work will expand its applications to other sectors and further assess its real-world effectiveness.
    \vspace{-3mm}
\section{Related Work}
Previous pFL methods for managing non-IID data issues typically fall into two categories: personalizing the global model and customizing individual models for each client.






   \vspace{-2mm}
\subsection{Global Model Personalization}

Global model personalization aims to adjust the global model to suit diverse client data distributions, creating a model that universally benefits all clients. This typically involves training the global model on varied data and local adaptations for each client's specific data. Previous studies have explored strategies to mitigate data heterogeneity and improve the global model’s generalization~\cite{pillutla2022federated,zhang2023gpfl}.


\vspace{-2mm}
\subsection{Learning Personalized Models}
Personalized model learning tailors individual models to each client's data, emphasizing local adaptation. This approach often employs weighted aggregation methods for local model personalization.

\vspace{-2mm}
\cparagraph{Model-wise aggregation.}
Train personalized models for each client by combining clients' models using weighted aggregation. For example, FedFomo \cite{zhang2021personalized} employs a distance metric for weighted aggregation, while APPLE \cite{luo2022adapt} introduces an adaptive mechanism to balance global and local objectives. FedAMP \cite{huang2021personalized} uses attention functions for client-specific models, and Ditto \cite{li2021ditto} incorporates a proximal term for personalized models. 
However, existing model aggregation methods may overlook complex variations and unique characteristics in client data, leading to suboptimal personalization.

\vspace{-2mm}
\cparagraph{Layer-wise aggregation.}
Customizes different layers to varying extents,  such as FedPer \cite{arivazhagan2019federated} and FedRep \cite{collins2021exploiting}. Moreover, pFedLA \cite{ma2022layerwised} uses hypernetworks to update layer-wise aggregation weights with a huge computation cost. All of them ignore the impact of diverse local data on the base and head layers during the training process, which limits further improvements in accuracy. 

\vspace{-2mm}
\cparagraph{Element-wise aggregation.}
This is the most fine-grained local aggregation approach, aggregating at the parameter level. FedALA \cite{zhang2023fedala} introduces an element-level aggregation weight matrix in the head layers, enhancing accuracy across various tasks. However, extra computation is required for weight calculation and does not account for the distinct roles and learning abilities of different layers during training.


\vspace{-3mm}
\section{Conclusion}

We have presented \SystemName, a new layer-wise pFL approach to optimize FL in the face of non-IID data. \SystemName adaptively adjusts the aggregation weights and learning rate and selects layer-wise masking to effectively incorporate local and global information throughout all network layers.
Experimental results show that \SystemName achieves the best inference accuracy and significantly reduces computational overheads compared to existing pFL methods.

\bibliography{aaai25}

\end{document}